\def\paperID{XXXX}
\def\confName{ICCV}
\def\confYear{2025}

\def\paperTitle{MDocAgent: A Multi-Modal Multi-Agent Framework for \\ Document Understanding}

\def\ours{MDocAgent}
\def\authorBlock{
    Siwei Han$^1$, Peng Xia$^1$, Ruiyi Zhang$^2$, Tong Sun$^2$, Yun Li$^1$, Hongtu Zhu$^1$, Huaxiu Yao$^1$ \\
    $^1$UNC-Chapel Hill, $^2$Adobe Research \\
    {\tt\small \{siweih,huaxiu\}@cs.unc.edu}
}

\newif\ifreview 
\newif\ifarxiv \newcommand{\arxiv}{\arxivtrue}
\newif\ifcamera 
\newif\ifrebuttal 

\arxiv 

\pdfoutput=1
\documentclass[10pt,twocolumn,letterpaper]{article}
\ifreview \usepackage[review]{cvpr} \fi
\ifarxiv \usepackage[pagenumbers]{cvpr} \fi
\ifrebuttal \usepackage[rebuttal]{cvpr} \fi
\ifcamera \usepackage{cvpr} \fi

\usepackage{algorithm}
\usepackage{algpseudocode}
\usepackage{graphicx}	
\usepackage{amsmath}	
\usepackage{amssymb}	
\usepackage{booktabs}
\usepackage{times}
\usepackage{microtype}
\usepackage{epsfig}
\usepackage{caption}
\usepackage{float}
\usepackage{placeins}
\usepackage{color, colortbl}
\usepackage{stfloats}
\usepackage{enumitem}
\usepackage{tabularx}
\usepackage{xstring}
\usepackage{multirow}
\usepackage{xspace}
\usepackage{url}
\usepackage{subcaption}
\usepackage{xcolor}
\usepackage{pifont}
\usepackage[hang,flushmargin]{footmisc}
\definecolor{myGreen}{RGB}{34, 139, 34}
\definecolor{myRed}{HTML}{FF6347}
\newcommand{\cmark}{\textcolor{myGreen}{\ding{51}}}
\newcommand{\xmark}{\textcolor{myRed}{\ding{55}}}
\ifcamera \usepackage[accsupp]{axessibility} \fi

% ...

\ifarxiv  \fi

\newcommand{\R}[1]{{%
    \textbf{%
        \ifstrequal{#1}{1}{\textcolor{red}{R#1}}{%
        \ifstrequal{#1}{2}{\textcolor{blue}{R#1}}{%
        \ifstrequal{#1}{3}{\textcolor{magenta}{R#1}}{%
        \ifstrequal{#1}{4}{\textcolor{teal}{R#1}}{%
                           \textcolor{cyan}{R#1}%
        }}}}%
    }%
}}
\usepackage{tcolorbox} 
\definecolor{graypurple}{RGB}{80, 60, 100} 
\definecolor{lightgraypurple}{RGB}{210, 200, 240} 

\tcbset{
    colback=lightgraypurple!20, 
    colframe=graypurple, 
    coltitle=white, 
    fonttitle=\bfseries, 
    arc=5mm, 
    boxrule=1pt, 
    sharp corners,
    before skip=5pt, 
    after skip=5pt 
}

\usepackage{xr-hyper}

\makeatletter
\newcommand*{\addFileDependency}[1]{
  \typeout{(#1)}
  \@addtofilelist{#1}
  \IfFileExists{#1}{}{\typeout{No file #1.}}
}

\makeatother
\newcommand*{\myexternaldocument}[1]{
    \externaldocument{#1}
    \addFileDependency{#1.tex}
    \addFileDependency{#1.aux}
}

\definecolor{cvprblue}{rgb}{0.21,0.49,0.74}
\usepackage[pagebackref,breaklinks,colorlinks,allcolors=cvprblue]{hyperref}
\usepackage[capitalize]{cleveref}
\crefname{section}{Sec.}{Secs.}
\crefname{table}{Table}{Tables}
\crefname{figure}{Fig.}{Figs.}

\ifarxiv \crefname{appendix}{App.}{Apps.}
\else \crefname{appendix}{Suppl.}{Suppls.} \fi

\unless\ifarxiv \myexternaldocument{_supplementary} \fi

\begin{document}
\title{\paperTitle}
\author{\authorBlock}
\maketitle

\begin{abstract}
Document Question Answering (DocQA) is a very common task. Existing methods using Large Language Models (LLMs) or Large Vision Language Models (LVLMs) and Retrieval Augmented Generation (RAG) often prioritize information from a single modal, failing to effectively integrate textual and visual cues. These approaches struggle with complex multi-modal reasoning, limiting their performance on real-world documents. We present \ours \ (A Multi-Modal Multi-Agent Framework for Document Understanding), a novel RAG and multi-agent framework that leverages both text and image. Our system employs five specialized agents: a general agent, a critical agent, a text agent, an image agent and a summarizing agent. These agents engage in multi-modal context retrieval, combining their individual insights to achieve a more comprehensive understanding of the document's content. This collaborative approach enables the system to synthesize information from both textual and visual components, leading to improved accuracy in question answering. Preliminary experiments on five benchmarks like MMLongBench, LongDocURL demonstrate the effectiveness of our \ours, achieve an average improvement of 12.1\% compared to current state-of-the-art method. This work contributes to the development of more robust and comprehensive DocQA systems capable of handling the complexities of real-world documents containing rich textual and visual information. Our data and code are available at \hyperlink{https://github.com/aiming-lab/MDocAgent}{https://github.com/aiming-lab/MDocAgent}.
\end{abstract}
\section{Introduction}
\label{sec:intro}

\begin{figure}[tp]
    \centering
    \includegraphics[width=0.95\textwidth, height=0.52\textheight, keepaspectratio]{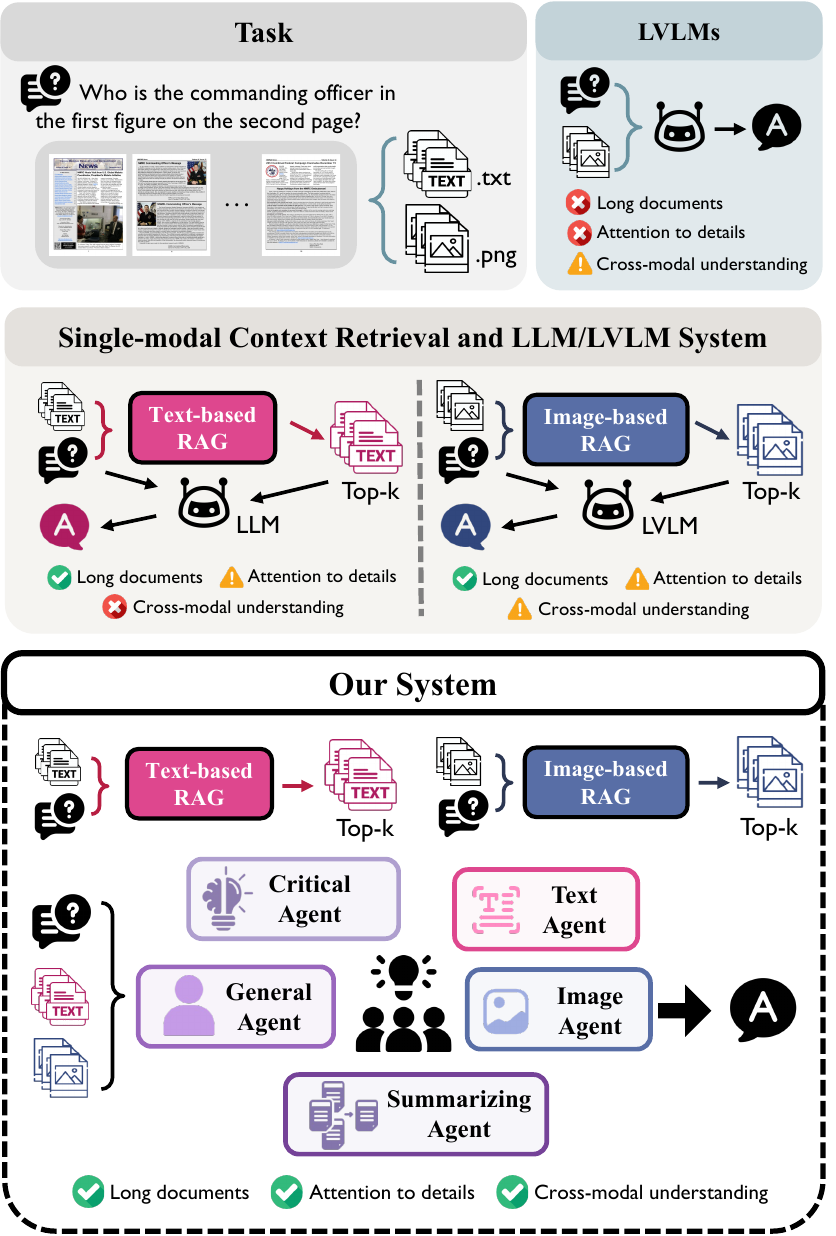}
    \caption{Comparison of different approaches for DocQA. LVLMs often struggle with long documents and lack granular attention to detail, while also exhibiting limitations in cross-modal understanding. Single-modal context retrieval can handle long documents but still suffers from issues with detailed analysis or integrating information across modalities. Our \ours\ addresses these challenges by combining text and image-based RAG with specialized agents for refined processing within each modality and a critical information extraction mechanism, showcasing improved DocQA performance.}
    \label{fig:intro_fig}
    \vspace{-2em}
\end{figure}
Answering questions based on reference documents (DocQA) is a critical task in many applications~\cite{ding2022v,tanaka2023slidevqa,mishra2019ocr,cho2024m3docrag,zhang2024ocr,ma2024visa,suri2024visdom}, ranging from information retrieval to automated document analysis. A key challenge in DocQA lies in the diverse nature of questions and the information needed to answer them~\cite{ma2024mmlongbenchdocbenchmarkinglongcontextdocument,deng2024longdocurl}. Questions can refer to textual content, to visual elements within the document (e.g., charts, diagrams, images), or even require the integration of information from both modalities.  Since Large Language Models (LLMs) can only handle textual information~\cite{naveed2023comprehensive}, Large Vision Language Models (LVLMs) are often used in DocQA~\cite{luo2024layoutllm,hu2024mplug,chen2024mllm}. As illustrated in Figure \ref{fig:intro_fig}, while LVLMs have shown promise in handling visual content, they often struggle in scenarios where key information is primarily textual, or where a nuanced understanding of the interplay between text and visual elements is required~\cite{cho2024m3docrag,ma2024visa,suri2024visdom}. Another challenge in DocQA lies in the huge volume of information often present in documents. Processing entire documents directly can overwhelm computational resources and make it difficult for models to identify the most pertinent information~\cite{ma2024mmlongbenchdocbenchmarkinglongcontextdocument,deng2024longdocurl}. 

To overcome this challenge, Retrieval Augmented Generation (RAG) is used as an auxiliary tool to extract the critical information from a long document~\cite{gao2023retrieval}. While RAG methods like ColBERT~\cite{khattab2020colbert} and ColPali~\cite{faysse2024colpali} have proven effective for retrieving textual or visual information respectively, they often fall short when a question requires integrating insights from both modalities. Existing RAG implementations typically operate in isolation, either retrieving text or images~\cite{lewis2020retrieval,xia2024mmed}, but lack the ability to synthesize information across these modalities. Consider a document containing a crucial diagram and accompanying textual explanations. If a question focuses on the diagram's content, a purely text-based RAG system would struggle to pinpoint the relevant information. Conversely, if the question pertains to a nuanced detail within the textual description, an image-based RAG would be unable to isolate the necessary textual segment. This inability to effectively combine multi-modal information restricts the performance of current RAG-based approaches in complex DocQA tasks. Moreover, the diverse and nuanced nature of these multimodal relationships requires not just retrieval, but also a mechanism for reasoning and drawing inferences across different modalities. 

To further address these limitations, we present a novel framework, a Multi-Modal Multi-Agent Framework for \\ Document Understanding (\ours), which leverages the power of both RAG and a collaborative multi-agent system where specialized agents collaborate to process and integrate text and image information. \ours\ employs two parallel RAG pipelines: a text-based RAG and an image-based RAG. These retrievers provide targeted textual and visual context for our multi-agent system. \ours\ comprises \textbf{five specialized agents}: a general agent for initial multi-modal processing, a critical agent for identifying key information, a text agent, an image agent for focused analysis within their respective modalities, and a summarizing agent to synthesize the final answer. This collaborative approach enables our system to effectively tackle questions that require synthesizing information from both textual and visual elements, going beyond the capabilities of traditional RAG methods.

\begin{figure*}[tp]
    \centering
    \includegraphics[width=0.95\textwidth, height=0.35\textheight, keepaspectratio]{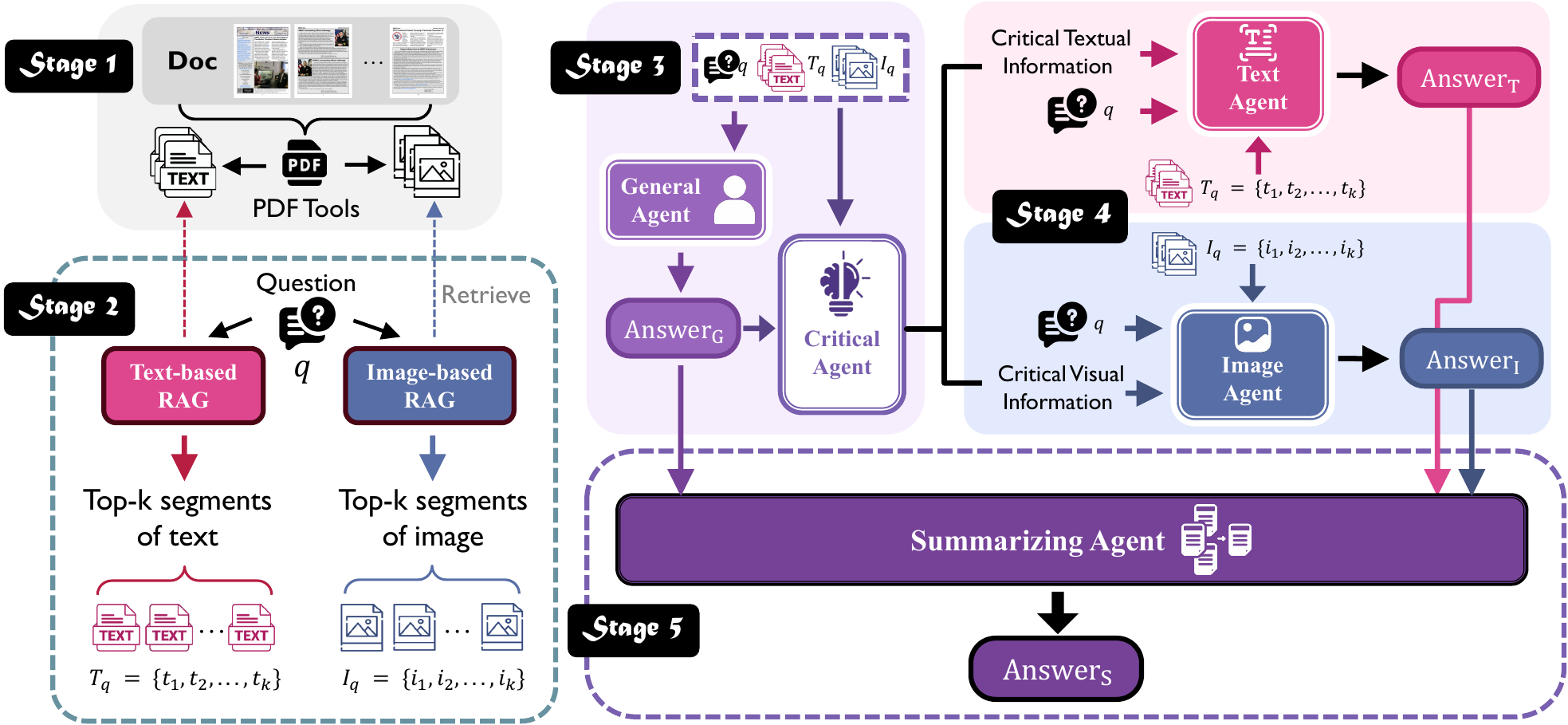}
    \caption{Overview of \textbf{\ours}: A multi-modal multi-agent framework operating in five stages: (1) Documents are processed using PDF tools to extract text and images. (2) Text-based and image-based RAG retrieves the top-k relevant segments and image pages. (3) The general agent provides a preliminary answer, and the critical agent extracts critical information from both modalities. (4) Specialized agents process the retrieved information and critical information within their respective modalities and generate refined answers. (5) The summarizing agent integrates all previous outputs to generate the final answer.}
    \label{fig:main_fig}
    \vspace{-1.5em}
\end{figure*}

Specifically, \ours\ operates in five stages: \textbf{(1) Document Pre-processing:} Text is extracted via OCR and pages are preserved as images. \textbf{(2) Multi-modal Context Retrieval:} text-based and image-based RAG tools retrieve the top-k relevant text segments and image pages, respectively. \textbf{(3) Initial Analysis and Key Extraction:} The general agent generates an initial response, and the critical agent extracts key information, providing it to the specialized agents. \textbf{(4) Specialized Agent Processing:} Text and image agents analyze the retrieved context within their respective modalities, guided by the critical information. \textbf{(5) Answer Synthesis:} The summarizing agent integrates all agent responses to produce the final answer.

The primary contribution of this paper is a novel multi-agent framework for DocQA that effectively integrates specialized agents, each dedicated to a specific modality or aspect of reasoning, including text and image understanding, critical information extraction, and answer synthesis. We demonstrate the efficacy of our approach through experiments on five benchmarks: MMLongBench~\cite{ma2024mmlongbenchdocbenchmarkinglongcontextdocument}, LongDocURL~\cite{deng2024longdocurl}, PaperTab~\cite{hui2024uda}, PaperText~\cite{hui2024uda}, and FetaTab~\cite{hui2024uda}, showing significant improvements in DocQA performance, with an average of 12.1\% compared to current SOTA method. The empirical improvements demonstrate the effectiveness of our collaborative multi-agent architecture in handling long, complex documents and questions. Furthermore, ablation studies validate the contribution of each agent and the importance of integrating multi-modalities.
\section{Related Work}
\label{sec:related}

\noindent \textbf{LVLMs in DocQA Tasks.} Document Visual Question Answering (DocVQA) has evolved from focusing on short documents to handling complex, long, and multi-document tasks~\cite{ding2022v,tanaka2023slidevqa,mishra2019ocr,tito2023hierarchical}, often involving visually rich content such as charts and tables. This shift requires models capable of integrating both textual and visual information. Large Vision Language Models (LVLMs) have emerged to address these challenges by combining the deep semantic understanding of Large Language Models (LLMs) with the ability to process document images~\cite{liu2023improved,liu2023visual,zhu2023minigpt,dai2023instructblip,zhou2023analyzing,zhou2024aligning,zhou2024calibrated,xia2024cares,xia2024mmie,zhu2024mmedpo,zhang2024motiongpt,tong2025mj}. LVLMs convert text in images into visual representations, preserving layout and visual context. However, they face challenges like input size limitations and potential loss of fine-grained textual details~\cite{luo2024layoutllm,hu2024mplug}, making effective integration of text and visual information crucial for accurate DocVQA performance~\cite{park2024hierarchical}.

\vspace{0.5em}
\noindent \textbf{Retrieval-Augmented Generation.} 
Retrieval Augmented Generation (RAG) enhances LLMs by supplying them with external text-based context, thereby improving their performance in tasks such as DocQA~\cite{lewis2020retrieval,gao2023retrieval}. Recently, with the increasing prevalence of visually rich documents, image RAG approaches have been developed to retrieve relevant visual content for Large Vision Language Models (LVLMs)~\cite{xia2024rule,xia2024mmed,cho2024m3docrag,chen2024mllm,xing2025re}. However, existing methods struggle to effectively integrate and reason over both text and image information, as retrieval often occurs independently. This lack of integrated reasoning limits the effectiveness of current RAG techniques, especially for complex DocQA tasks that require a nuanced understanding of both modalities.

\vspace{0.5em}
\noindent \textbf{Multi-Agent Systems.}
Multi-agent systems have shown promise in complex domains like medicine~\cite{wu2023autogen,li2023camel,kim2024mdagents}. These systems use specialized agents to focus on different task aspects~\cite{su2020adapting,chan2023chateval,kannan2024smart,li2025metal}, collaborating to achieve goals that a single model may struggle with. However, their application to DocQA introduces unique challenges stemming from the need to integrate diverse modalities. Simply combining the outputs of independent text and image agents often fails to capture the nuanced interplay between these modalities, which is crucial for accurate document understanding. Our framework addresses this by introducing \textit{a general agent for information integration} alongside specialized text and image agents, enabling collaborative reasoning and a more comprehensive understanding of document content, ultimately improving DocVQA performance.

\section{Multi-Modal Multi-Agent Framework for Document Understanding}
\label{sec:method}
This section details our proposed framework, \ours, for tackling the complex challenges of DocQA. \ours\ employs a novel five-stage multi-modal, multi-agent approach as shown in Figure \ref{fig:main_fig}, utilizing specialized agents for targeted information extraction and cross-modal synthesis to achieve a more comprehensive understanding of document content. Subsequently, Section 3.1 through Section 3.5 provide a comprehensive description of \ours's architecture. This detailed exposition will elucidate the mechanisms by which \ours\ effectively integrates and leverages textual and visual information to achieve improved accuracy in DocQA.

\noindent \textbf{Preliminary: Document Question Answering.} Given a question $q$ expressed in natural language and the corresponding document $\mathcal{D}$, the goal is to generate an answer a that accurately and comprehensively addresses $q$ using the information provided within $\mathcal{D}$. 

\subsection{Document Pre-Processing}
This initial stage prepares the document corpus for subsequent processing by transforming it into a format suitable for both textual and visual analysis. $\mathcal{D}$ consists of a set of pages $\mathcal{D} = \{p_{1}, p_{2}, \ldots, p_{N}\}$. For each page $p_{i}$, textual content is extracted using a combination of Optical Character Recognition (OCR) and PDF parsing techniques. OCR is employed to recognize text within image-based PDFs, while PDF parsing extracts text directly from digitally encoded text within the PDF. This dual approach ensures robust text extraction across various document formats and structures. The extracted text for each page $p_{i}$ is represented as a sequence of textual segments or paragraphs $t_{i} = \{t_{i1}, t_{i2}, \ldots, t_{iM}\}$, where $M$ represents the number of text segments on that page. Concurrently, each page $p_{i}$ is also preserved as an image, retaining its original visual layout and features. This allows the framework to leverage both textual and visual cues for comprehensive understanding. This pre-processing results in two parallel representations of the document corpus: a textual representation consisting of extracted text segments and a visual representation consisting of the original page images. This dual representation forms the foundation for the multi-modal analysis performed by the framework.
    
\subsection{Multi-modal Context Retrieval} The second stage focuses on efficiently retrieving the most relevant information from the document corpus, considering both text and image modalities. Algorithm \ref{alg:multi-modal_context_retrieval} illustrates the whole procedure of retrieval. For the textual retrieval, extracted text segments $t_{i}$ of each page $p_{i}$ are indexed using ColBERT~\cite{khattab2020colbert}. Given the user question $q$, ColBERT retrieves the top-$k$ most relevant text segments, denoted as $T_q = \{t_1, t_2, \ldots, t_k\}$. This provides the textual context for subsequent agent processing. Parallel to textual retrieval, visual context is extracted using ColPali~\cite{faysse2024colpali}. Each page image $p_{i}$ is processed by ColPali to generate a dense visual embedding $E^{p_{i}} \in \mathbb{R}^{n^v \times d}$, where $n^v$ represents the number of visual tokens per page and $d$ represents the embedding dimension. Using these embeddings and the question $q$, ColPali retrieves the top-$k$ most visually relevant pages, denoted as $I_q = \{i_1, i_2, \ldots, i_k\}$. The use of ColPali allows the model to capture the visual information present in the document, including layout, figures, and other visual cues.

\begin{algorithm}
\caption{Multi-modal Context Retrieval}
\label{alg:multi-modal_context_retrieval}
\begin{algorithmic}[1]
\Require Question $q$, Document $D$, Text Scores $S_t$, Image Scores $S_i$, Text Relevance Scores $R_t$, Image Relevance Scores $R_i$.
\Ensure Top-k text segments $T_q$, Top-k image segments $T_q$.

\State $S_t \gets \{\}$
\State $S_i \gets \{\}$
\Comment{\textcolor{blue}{Iterate through each page in the corpus}}
\For{each $p$ in $D$} 
    \For{each text segment $t$ in $p$} 
        \State $S_t[t] \gets R_t(q, t)$\Comment{\textcolor{blue}{Calculate text relevance score}}
    \EndFor
    \State $S_i[p] \gets R_i(q, p)$ \Comment{\textcolor{blue}{Calculate image relevance score}}
\EndFor

\State $T_q \gets \text{Top\_K}(S_t, k)$ \Comment{\textcolor{blue}{Select top-k text segments}}
\State $I_q \gets \text{Top\_K}(S_i, k)$ \Comment{\textcolor{blue}{Select top-k image segments}}

\State \Return $T_q$, $I_q$

\end{algorithmic}
\end{algorithm}

\subsection{Initial Analysis and Key Extraction} The third stage aims to provide an initial interpretation of the question and pinpoint the most salient information within the retrieved context. The general agent $A_G$, functioning as a preliminary multi-modal integrator, receives both the retrieved textual context $T_q$ and the visual context $I_q$. It processes these multimodal inputs by effectively combining the information embedded within both modalities. This comprehensive understanding of the combined context allows $A_G$ to generate a preliminary answer $a_G$, which serves as a crucial starting point for more specialized analysis in the next stage. 
\begin{equation}
    a_G = A_G(q, T_q, I_q).
\end{equation}
Subsequently, the critical agent $A_C$ plays a vital role in refining the retrieved information. It takes as input the question $q$, the retrieved contexts $T_q$ and $I_q$, and the preliminary answer $a_G$ generated by the general agent. The primary function of $A_C$ is to meticulously analyze these inputs and identify the most crucial pieces of information that are essential to accurately answer the question. This critical information acts as a guide for the specialized agents in the next stage, focusing their attention on the most relevant aspects of the retrieved context. 
\begin{equation}
T_c = A_C(q, T_q, a_G), \quad
I_c = A_C(q, I_q, a_G).
\end{equation}
The output of this stage consists of $T_c \subset T_q$, representing the critical textual information extracted from the retrieved text segments, and $I_c$, which provides a detailed textual description of the critical visual information extracted from the retrieved images $I_q$ that capture the essence of the important visual elements.

\subsection{Specialized Agent Processing} 
The fourth stage delves deeper into the textual and visual modalities, leveraging specialized agents guided by the critical information extracted in the previous stage. The text agent $A_T$ receives the retrieved text segments $T_q$ and the critical textual information $T_c$ as input. It operates exclusively within the textual domain, leveraging its specialized knowledge and analytical capabilities to thoroughly examine the provided text segments. By focusing specifically on the critical textual information $T_c$, $A_T$ can pinpoint the most relevant evidence within the broader textual context $T_q$ and perform a more focused analysis. This focused approach allows for a deeper understanding of the textual nuances related to the question and culminates in the generation of a detailed, text-based answer $a_T$. 
\begin{equation}
    a_T = A_T(q, T_q, T_c).
\end{equation}
Concurrently, the image agent $A_I$ receives the retrieved images $I_q$ and the critical visual information $I_c$. This agent specializes in visual analysis and interpretation. It processes the images in $I_q$, paying particular attention to the regions or features highlighted by the critical visual information $I_c$. This targeted analysis allows the agent to extract valuable insights from the visual content, focusing its processing on the most relevant aspects of the images. The image agent's analysis results in a visually-grounded answer $a_I$, which provides a response based on the interpretation of the images.
\begin{equation}
    a_I = A_I(q, I_q, I_c).
\end{equation}

\subsection{Answer Synthesis} The final stage integrates the diverse outputs from the preceding stages, combining the initial multi-modal understanding with the specialized agent analyses to produce a comprehensive and accurate answer. The summarizing agent $A_S$ receives the answers $a_G$, $a_T$, and $a_I$ generated by the general agent, text agent, and image agent, respectively. This comprehensive set of information provides a multifaceted perspective on the question and allows the summarizing agent to perform a thorough synthesis. The summarizing agent analyzes the individual agent answers, identifying commonalities, discrepancies, and complementary insights. It considers the supporting evidence provided by each agent. By resolving potential conflicts or disagreements between the agents and integrating their individual strengths, the summarizing agent constructs a final answer $a_S$ that leverages the collective intelligence of the multi-agent system. This final answer is not merely a combination of individual answers but a synthesized response that reflects a deeper and more nuanced understanding of the information extracted from both textual and visual modalities. The whole procedure of this multi-agent collaboration is illustrated in Algorithm \ref{alg:multi-agent_processing}.

\begin{algorithm}
\caption{Multi-agent Collaboration}
\label{alg:multi-agent_processing}
\begin{algorithmic}[1]
\Require Question $q$, Top-k text segments $T_q$, Top-k image segments $I_q$, General Agent $A_G$, Critical Agent $A_C$, Text Agent $A_T$, Image Agent $A_I$, Summarizing Agent $A_S$
\Ensure Final answer $a_s$,

\State $a_G \gets A_G(q, T_q, I_q)$ \Comment{\textcolor{blue}{General agent answer}}
\State $(T_c, B_c) \gets A_C(q, T_q, I_q, a_G)$ \Comment{\textcolor{blue}{Extract critical info}}

\State $a_T \gets A_T(q, T_q, T_c)$ \Comment{\textcolor{blue}{Text agent answer}}
\State $a_I \gets A_I(q, I_q, B_c)$ \Comment{\textcolor{blue}{Image agent answer}}

\State $a_S \gets A_S(q, a_G, a_T, a_I)$ \Comment{\textcolor{blue}{Final answer synthesis}}

\State \Return $a_S$
\end{algorithmic}
\end{algorithm}

\vspace{-1em}
\begin{table*}[t]
    \centering
    \renewcommand{\arraystretch}{1.2}
    \small
        \caption{Performance comparison across \ours\ and existing state-of-the-art LVLMs and RAG-based methods.}

    \resizebox{0.95\textwidth}{!}{
    \setlength{\tabcolsep}{6pt}
    \begin{tabular}{l|cccccc}
        \toprule
        \textbf{Method} & \textbf{MMLongBench} & \textbf{LongDocUrl} & \textbf{PaperTab} & \textbf{PaperText} & \textbf{FetaTab} & \textbf{Avg} \\
        \midrule
        \multicolumn{7}{c}{\textit{LVLMs}} \\
        \midrule
        Qwen2-VL-7B-Instruct~\cite{Qwen2-VL} & 0.165 & 0.296 & 0.087 & 0.166 & 0.324 & 0.208 \\
        Qwen2.5-VL-7B-Instruct~\cite{Qwen2.5-VL} & 0.224 & 0.389 & 0.127 & 0.271 & 0.329 & 0.268 \\
        LLaVA-v1.6-Mistral-7B~\cite{liu2023improved} & 0.099 & 0.074 & 0.033 & 0.033 & 0.110 & 0.070 \\
        Phi-3.5-Vision-Instruct~\cite{abdin2024phi} & 0.144 & 0.280 & 0.071 & 0.165 & 0.237 & 0.179 \\
        LLaVA-One-Vision-7B~\cite{li2024llava} & 0.053 & 0.126 & 0.056 & 0.108 & 0.077 & 0.084 \\
        SmolVLM-Instruct~\cite{marafioti2025smolvlm} & 0.081 & 0.163 & 0.066 & 0.137 & 0.142 & 0.118\\
        \midrule
        \multicolumn{7}{c}{\textit{RAG methods (top 1)}} \\
        \midrule
        ColBERTv2~\cite{santhanam2021colbertv2}+LLaMA-3.1-8B~\cite{grattafiori2024llama} & 0.241 & 0.429 & 0.155 & 0.332 & 0.490 & 0.329 \\
        M3DocRAG~\cite{cho2024m3docrag} (ColPali~\cite{faysse2024colpali}+Qwen2-VL-7B~\cite{Qwen2-VL}) & 0.276 & 0.506 & 0.196 & 0.342 & 0.497 & 0.363 \\
        \textbf{\ours\ (Ours)} & \textbf{0.299} & \textbf{0.517} & \textbf{0.219} & \textbf{0.399} & \textbf{0.600} & \textbf{0.407} \\
        \midrule
        \multicolumn{7}{c}{\textit{RAG methods (top 4)}} \\
        \midrule
        ColBERTv2~\cite{santhanam2021colbertv2}+LLaMA-3.1-8B~\cite{grattafiori2024llama} & 0.273 & 0.491 & 0.277 & 0.460 & 0.673 & 0.435 \\
        M3DocRAG~\cite{cho2024m3docrag} (ColPali~\cite{faysse2024colpali}+Qwen2-VL-7B~\cite{Qwen2-VL}) & 0.296 & 0.554 & 0.237 & 0.430 & 0.578 & 0.419 \\
        \textbf{\ours\ (Ours)} & \textbf{0.315} & \textbf{0.578} & \textbf{0.278} & \textbf{0.487} & \textbf{0.675} & \textbf{0.465} \\
        \bottomrule
    \end{tabular}
    }
    \label{tab:main-results}
    \vspace{-0.2em}
\end{table*}

\begin{table*}[ht]
    \centering
    \caption{Performance comparison across different \ours's variants.}
    \renewcommand{\arraystretch}{1.2}
    \resizebox{1\textwidth}{!}{
    \setlength{\tabcolsep}{8pt}
    \begin{tabular}{c|ccc|cccccc}
        \toprule
        \multirow{2}{*}{\textbf{Variants}} & \multicolumn{3}{c|}{\textbf{Agent Configuration}} & \multicolumn{6}{c}{\textbf{Evaluation Benchmarks}} \\  
        \cmidrule(lr){2-4}
        \cmidrule(lr){5-10}
        & {\textbf{General \& Critical Agent}} & {\textbf{Text Agent}} & {\textbf{Image Agent}} & \textbf{MMLongBench} & \textbf{LongDocUrl} & \textbf{PaperTab} & \textbf{PaperText} & \textbf{FetaTab} & \textbf{Avg} \\
        
        \midrule
        \textbf{\ours$_i$} & \cmark & \xmark & \cmark & 0.287 & 0.508 & 0.196 & 0.376 & 0.552 & 0.384 \\
        \textbf{\ours$_t$} & \cmark & \cmark & \xmark & 0.288 & 0.484 & 0.201 & 0.391 & 0.596 & 0.392 \\
        \textbf{\ours$_s$} & \xmark & \cmark & \cmark & 0.285 & 0.479 & 0.188 & 0.365 & 0.592 & 0.382 \\\midrule
        \textbf{\ours} & \cmark & \cmark & \cmark & \textbf{0.299} & \textbf{0.517} & \textbf{0.219} & \textbf{0.399} & \textbf{0.600} & \textbf{0.407} \\
        \bottomrule
    \end{tabular}
    }
    \label{tab:ablation1}
\end{table*}

\section{Experiments}
\label{sec:experiments}
We evaluate \ours\ on five document understanding benchmarks covering multiple scenarios to answer the following questions: (1) Does \ours\ effectively improve document understanding accuracy compared to existing RAG-based approaches? (2) Does each agent in our framework play a meaningful role? (3) How does our approach enhance the model's understanding of documents?
\subsection{Experiment Setup}
\textbf{Implementation Details}. There are five agents in \ours: general agent, critical agent, text agent, image agent and summarizing agent. We adopt Llama-3.1-8B-Instruct~\cite{grattafiori2024llama} as the base model for text agent, Qwen2-VL-7B-Instruct~\cite{Qwen2-VL} for other four agents, and select ColBERTv2~\cite{santhanam2021colbertv2} and ColPali~\cite{faysse2024colpaliefficientdocumentretrieval} as the text and image retrievers, respectively. In our settings of RAG, we retrieve 1 or 4 highest-scored segments as input context for each example. All experiments are conducted on 4 NVIDIA H100 GPUs. Details of models and settings are shown in Appendix \ref{sec:appendix_section}.

\noindent
\textbf{Datasets}. The benchmarks involve MMLongBench~\cite{ma2024mmlongbenchdocbenchmarkinglongcontextdocument}, LongDocUrl~\cite{deng2024longdocurl}, PaperTab~\cite{hui2024uda}, PaperText~\cite{hui2024uda}, FetaTab~\cite{hui2024uda}. These evaluation datasets cover a variety of scenarios, including both open- and closed-domain, textual and visual, long and short documents, ensuring fairness and completeness in the evaluation. Details of dataset descriptions are in Appendix \ref{sec:appendix_datainfo}.

\noindent
\textbf{Metrics}. For all benchmarks, following~\citet{ma2024mmlongbenchdocbenchmarkinglongcontextdocument,deng2024longdocurl}, we leverage GPT-4o~\cite{openai2023gpt4} as the evaluation model to assess the consistency between the model's output and the reference answer, producing a binary decision (correct/incorrect). We provide the average accuracy rate for each benchmark.

\subsection{Main Results}
In this section, we provide a comprehensive comparison of \ours\ on multiple benchmarks against existing state-of-the-art LVLMs and RAG-based methods built on them. Our findings can be summarized as:

\begin{table*}[htbp]
\centering
\caption{Performance comparison across different evidence source on MMLongBench.}
\vspace{-0.5em}
\small
\renewcommand{\arraystretch}{1.2}
\setlength{\tabcolsep}{5pt}
\begin{tabular}{lcccccc}
\toprule
\textbf{Method} & \textbf{Chart} & \textbf{Table} & \textbf{Pure-text} & \textbf{Generalized-text} & \textbf{Figure} & \textbf{Avg} \\
\midrule
\multicolumn{7}{c}{\textit{LVLMs (up to 32 pages)}} \\
\midrule
Qwen2-VL-7B-Instruct & 0.182 & 0.097 & 0.209 & 0.185 & 0.197 & 0.165 \\
Qwen2.5-VL-7B-Instruct & 0.188 & 0.124 & 0.265 & 0.210 & 0.254 & 0.224 \\
LLaVA-v1.6-Mistral-7B & 0.011 & 0.023 & 0.033 & 0.000 & 0.057 & 0.074 \\
LLaVA-One-Vision-7B & 0.045 & 0.051 & 0.076 & 0.017 & 0.084 & 0.053 \\
Phi-3.5-Vision-Instruct & 0.159 & 0.101 & 0.156 & 0.160 & 0.164 & 0.144 \\
SmolVLM-Instruct & 0.062 & 0.065 & 0.123 & 0.118 & 0.094 & 0.081 \\
\midrule
\multicolumn{7}{c}{\textit{RAG methods (top 1)}} \\
\midrule
ColBERTv2+LLaMA-3.1-8B & 0.148 & 0.203 & 0.265 & 0.143 & 0.074 & 0.241 \\
M3DocRAG (ColPali+Qwen2-VL-7B) & 0.268 & 0.263 & 0.334 & 0.250 & \textbf{0.303} & 0.276 \\
\textbf{\ours\ (Ours)} & \textbf{0.269} & \textbf{0.300} & \textbf{0.348} & \textbf{0.252} & 0.298 & \textbf{0.299} \\
\midrule
\multicolumn{7}{c}{\textit{RAG methods (top 4)}} \\
\midrule
ColBERTv2+LLaMA-3.1-8B & 0.182 & 0.267 & 0.311 & 0.168 & 0.120 & 0.273 \\
M3DocRAG (ColPali+Qwen2-VL-7B) & 0.290 & 0.318 & 0.371 & 0.277 & \textbf{0.321} & 0.296 \\
\textbf{\ours\ (Ours)} & \textbf{0.347} & \textbf{0.323} & \textbf{0.401} & \textbf{0.294} & \textbf{0.321} & \textbf{0.315} \\
\bottomrule
\end{tabular}
\label{tab:mmlb-results}
\end{table*}

\begin{table*}[ht]
    \centering
        \caption{Performance comparison between using ColPali and ColQwen2-v1.0 as \ours's image-based RAG model.}
    \renewcommand{\arraystretch}{1.2}
    \resizebox{0.75\textwidth}{!}{
    \setlength{\tabcolsep}{8pt}
    \begin{tabular}{l|cccccc}
        \toprule
        & \textbf{MMLongBench} & \textbf{LongDocUrl} & \textbf{PaperTab} & \textbf{PaperText} & \textbf{FetaTab} & \textbf{Avg} \\
        \midrule
        \textbf{+ColPali} & 0.299 & 0.517 & \textbf{0.219} & \textbf{0.399} & 0.600 & \textbf{0.407} \\
        \textbf{+ColQwen2-v1.0} & \textbf{0.303} & \textbf{0.520} & 0.216 & 0.391 & \textbf{0.603} & \textbf{0.407} \\
        \bottomrule
    \end{tabular}
    }
    \label{tab:ablation3}
    \vspace{-1em}
\end{table*}

\noindent
\textbf{\ours\ Outperforms All the Comparison Methods and Other LVLMs}. We compare our method with baseline approaches on document understanding tasks, with the results presented in Table \ref{tab:main-results}. Overall, our method outperforms all baselines across all benchmarks. 

\noindent
\textbf{Top-1 Retrieval Performance.} With top-1 retrieval, \ours\ demonstrates a significant performance improvement. On PaperText, \ours\ achieves a score of 0.399, surpassing the second-best method, M3DocRAG, by 16.7\%. Similarly, on FetaTab, \ours\ attains a score of 0.600, exceeding the second-best method by an impressive 21.0\%. Compared to the best LVLM (Qwen2.5-VL-7B) and text-RAG-based (ColBERTv2+Llama-3.1-8B) baselines, our approach demonstrates a remarkable average improvement of 51.9\% and 23.7\% on average across all benchmarks. This improvement highlights the benefits of incorporating visual information and the collaborative multi-agent architecture in our framework. Furthermore, recent state-of-the-art image-RAG-based method M3DocRAG~\cite{cho2024m3docrag} show promising results, yet our approach still outperforms it by 12.1\% on average. 
This suggests that our multi-agent framework, with its specialized agents and critical information extraction mechanism addresses the core challenges of information overload, granular attention to detail, and cross-modality understanding more effectively than existing methods.

\noindent \textbf{Top-4 Retrieval Performance.} When using top-4 retrieval, the advantages of our method are further demonstrated. \ours\ consistently achieves the highest scores across all benchmarks. On average, \ours\ outperforms Qwen2.5-VL-7B by a remarkable 73.5\%. Interestingly, with top-4 retrieval, M3DocRAG slightly performs worse than ColBERTv2+Llama-3.1-8B compared to top-1 retrieval. This may suggest limitations on M3DocRAG's capacity of selectively integrate across multiple retrieved documents when dealing with larger amounts of retrieved information. On average, \ours\ exceeds M3DocRAG by 10.9\%. Meanwhile, compared to ColBERTv2+Llama-3.1-8B, \ours\ demonstrates a 6.9\% improvement. This consistent improvement suggests that our method effectively harnesses the additional contextual information provided by the top-4 retrieved items, offering a greater benefit with more retrieval results.

\subsection{Quantitative Analysis}
In this section, we conduct three quantitative analyses to understand the effectiveness and contribution of different components within our proposed framework. First, we perform ablation studies to assess the impact of removing individual agents or groups of agents. Second, we present a fine-grained performance analysis, examining \ours's performance across different evidence modalities on MMLongBench to pinpoint the source of its improvements. Third, a compatibility analysis explores the framework's performance with different image-based RAG backbones to demonstrate its robustness and generalizability. Additionally, we present experimental results showcasing its performance with different model backbones in Appendix \ref{sec:appendix_more_lvlms}.

\noindent
\subsubsection{Ablation Studies}
Table \ref{tab:ablation1} presents a comparison of our full method (\ours) against it's variants: \ours$_i$ (without the text agent) and \ours$_t$ (without the image agent). Across all benchmarks, the full \ours\ method consistently achieves the highest performance. The removal of either specialized agent, text or image, results in a noticeable performance drop. This underscores the importance of incorporating both text and image modalities through specialized agents within our framework. The performance difference is most pronounced in benchmarks like LongDocURL and PaperText, which likely contain richer visual or textual information respectively, further highlighting the value of specialized processing. This ablation study clearly demonstrates the synergistic effect of combining specialized agents dedicated to each modality.

Table \ref{tab:ablation1} also compares \ours\ with \ours$_s$, where both the general agent and the critical agent are removed, to evaluate their contribution. The consistent improvement of the full method over \ours$_s$ across all datasets clearly underscores the importance of these two agents. The general agent establishes a crucial foundation by initially integrating both text and image modalities, providing a holistic understanding of the context. Removing this integration step noticeably reduces the subsequent agents' capacity to focus their analysis of critical information and answer effectively. On top of general modal integration, removing the critical agent limits the framework's ability to effectively identify and leverage crucial information. This highlights the essential role of the critical agent in focusing the specialized agents' attention and facilitating more targeted and efficient information extraction.

\subsubsection{Fine-Grained Performance Analysis}
We present an in-depth analysis of the performance in different types of evidence modalities, by further analyzing the scores on MMLongBench in Table \ref{tab:mmlb-results}, to gain a better understanding of the performance improvements achieved by \ours. We also illustrate the results of evidence modalities of LongDocURL in Appendix \ref{sec:appendix_ldu}. According to the results, \ours\ outperforms all LVLM baselines among all types of evidence modalities. When comparing RAG methods using the top 1 retrieval approach, though M3DocRAG performs slightly better on Figure category, \ours\ show strong performance in Chart, Table and Text categories, reflecting its enhanced capability to process textual and visual information. With the top 4 retrieval strategy, \ours\ enhances its performance in the all categories, specifically in Figure, highlighting its effective handling of large and varied information sources.

\subsubsection{Compatibility Analysis}
We further analyze the compatibility of \ours\ with different RAG backbones. Table \ref{tab:ablation3} presents results using two image-based RAG models, ColPali and ColQwen2-v1.0, within our proposed framework. Both models achieve comparable overall performance, with an identical average score of 0.407 across all benchmarks. While ColQwen2-v1.0 shows a slight advantage on MMLongBench, LongDocUrl, and FetaTab, ColPali performs marginally better on PaperTab and PaperText. This suggests that the choice of image-based RAG model has minimal impact on the framework’s overall effectiveness, underscoring the robustness of our multi-agent architecture. Moreover, the consistency in performance across different RAG models highlights that the core strength of our approach lies in the multi-agent architecture itself, rather than reliance on a specific retrieval model. This further reinforces the compatibility of our proposed method.

\begin{figure*}[tp]
    \centering
    \includegraphics[width=1\textwidth, height=0.4\textheight, keepaspectratio]{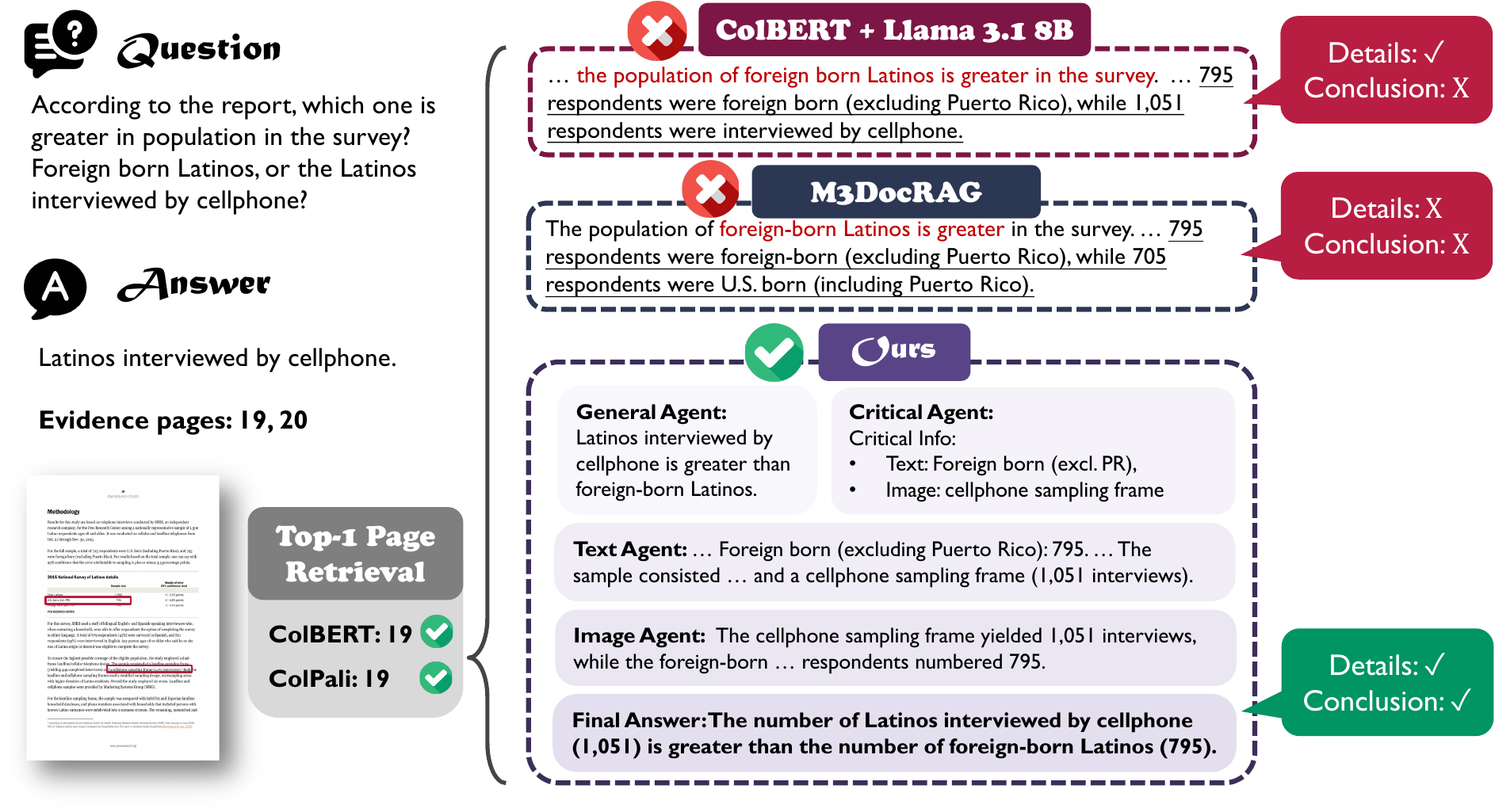}
    \caption{A Case study of \textbf{\ours} compared with other two RAG-method baselines(ColBERT + Llama 3.1-8B and M3DocRAG). Given a question comparing two population sizes, both baseline methods  fail to arrive at the correct answer. Our framework, through the collaborative efforts of its specialized agents, successfully identifies the relevant information from both text and a table within the image, ultimately synthesizing the correct answer. This highlights the importance of granular, multi-modal analysis and the ability to accurately process information within the context.}
    \label{fig:case_study}
    \vspace{-0.5em}
\end{figure*}

\subsection{Case Study}

We perform a case study to better understand \ours. Figure 3 illustrates an example. The question requires extracting and comparing numerical information related to two distinct Latino populations from both textual and tabular data within a document. While both ColBERT and ColPali successfully retrieve the relevant page containing the necessary information, both baseline methods fail to synthesize the correct answer. The ColBERT + Llama-3.1-8B baseline, relying solely on text, incorrectly concludes that the foreign-born Latino population is greater, demonstrating a failure to accurately interpret the numerical data presented within the document's textual content. Similarly, M3DocRAG fails to correctly interpret the question due to capturing wrong information. In contrast, our multi-agent framework successfully navigates this complexity and gives the correct answer. 

Specifically, the general agent provides a correct but vague answer, making the critical agent essential for identifying key phrases like ``Foreign born (excl. PR)" and the ``cellphone sampling frame" table. This guides specialized agents to precise locations for efficient data extraction. Both text agent and image agent correctly extract 795 for foreign-born Latinos and 1,051 for cellphone-interviewed Latinos. The summarizing agent then integrates these insights for accurate comparison and a comprehensive final answer. This case study demonstrates how our structured, multi-agent framework outperforms methods struggling with integrated text and image analysis (See more case studies in Appendix~\ref{sec:appendix_cases}).
\section{Conclusion}
\label{sec:conclusion}

This paper presents a multi-agent framework \ours\ for DocQA that integrates text and visual information through specialized agents and a dual RAG approach. Our framework addresses the limitations of existing methods by employing agents dedicated to text processing, image analysis, and critical information extraction, culminating in a synthesizing agent for final answer generation. Experimental results demonstrate significant improvements over LVLMs and multi-modal RAG methods, highlighting the efficacy of our collaborative multi-agent architecture. Our framework effectively handles information overload and promotes detailed cross-modal understanding, leading to more accurate and comprehensive answers in complex DocQA tasks. Future work will explore more advanced inter-agent communication and the integration of external knowledge sources.

\section*{Acknowledgement}
This research was partially supported by NIH 1R01AG085581 and Cisco Faculty Research Award.
{\small
\bibliographystyle{ieeenat_fullname}
\bibliography{11_references}

\begin{thebibliography}{51}
\providecommand{\natexlab}[1]{#1}
\providecommand{\url}[1]{\texttt{#1}}
\expandafter\ifx\csname urlstyle\endcsname\relax
  \providecommand{\doi}[1]{doi: #1}\else
  \providecommand{\doi}{doi: \begingroup \urlstyle{rm}\Url}\fi

\bibitem[Abdin et~al.(2024)Abdin, Aneja, Awadalla, Awadallah, Awan, Bach, Bahree, Bakhtiari, Bao, Behl, et~al.]{abdin2024phi}
Marah Abdin, Jyoti Aneja, Hany Awadalla, Ahmed Awadallah, Ammar~Ahmad Awan, Nguyen Bach, Amit Bahree, Arash Bakhtiari, Jianmin Bao, Harkirat Behl, et~al.
\newblock Phi-3 technical report: A highly capable language model locally on your phone.
\newblock \emph{arXiv preprint arXiv:2404.14219}, 2024.

\bibitem[Bai et~al.(2025)Bai, Chen, Liu, Wang, Ge, Song, Dang, Wang, Wang, Tang, Zhong, Zhu, Yang, Li, Wan, Wang, Ding, Fu, Xu, Ye, Zhang, Xie, Cheng, Zhang, Yang, Xu, and Lin]{Qwen2.5-VL}
Shuai Bai, Keqin Chen, Xuejing Liu, Jialin Wang, Wenbin Ge, Sibo Song, Kai Dang, Peng Wang, Shijie Wang, Jun Tang, Humen Zhong, Yuanzhi Zhu, Mingkun Yang, Zhaohai Li, Jianqiang Wan, Pengfei Wang, Wei Ding, Zheren Fu, Yiheng Xu, Jiabo Ye, Xi Zhang, Tianbao Xie, Zesen Cheng, Hang Zhang, Zhibo Yang, Haiyang Xu, and Junyang Lin.
\newblock Qwen2.5-vl technical report.
\newblock \emph{arXiv preprint arXiv:2502.13923}, 2025.

\bibitem[Chan et~al.(2023)Chan, Chen, Su, Yu, Xue, Zhang, Fu, and Liu]{chan2023chateval}
Chi-Min Chan, Weize Chen, Yusheng Su, Jianxuan Yu, Wei Xue, Shanghang Zhang, Jie Fu, and Zhiyuan Liu.
\newblock Chateval: Towards better llm-based evaluators through multi-agent debate.
\newblock \emph{arXiv preprint arXiv:2308.07201}, 2023.

\bibitem[Chen et~al.(2024)Chen, Xu, Qi, and Guo]{chen2024mllm}
Zhanpeng Chen, Chengjin Xu, Yiyan Qi, and Jian Guo.
\newblock Mllm is a strong reranker: Advancing multimodal retrieval-augmented generation via knowledge-enhanced reranking and noise-injected training.
\newblock \emph{arXiv preprint arXiv:2407.21439}, 2024.

\bibitem[Cho et~al.(2024)Cho, Mahata, Irsoy, He, and Bansal]{cho2024m3docrag}
Jaemin Cho, Debanjan Mahata, Ozan Irsoy, Yujie He, and Mohit Bansal.
\newblock M3docrag: Multi-modal retrieval is what you need for multi-page multi-document understanding.
\newblock \emph{arXiv preprint arXiv:2411.04952}, 2024.

\bibitem[Dai et~al.(2023)Dai, Li, Li, Tiong, Zhao, Wang, Li, Fung, and Hoi]{dai2023instructblip}
Wenliang Dai, Junnan Li, Dongxu Li, Anthony Meng~Huat Tiong, Junqi Zhao, Weisheng Wang, Boyang Li, Pascale Fung, and Steven Hoi.
\newblock Instructblip: Towards general-purpose vision-language models with instruction tuning.
\newblock \emph{arXiv preprint arXiv:2305.06500}, 2023.

\bibitem[Deng et~al.(2024)Deng, Yuan, Bu, Wang, Li, Xu, Li, Gao, Song, Zheng, et~al.]{deng2024longdocurl}
Chao Deng, Jiale Yuan, Pi Bu, Peijie Wang, Zhong-Zhi Li, Jian Xu, Xiao-Hui Li, Yuan Gao, Jun Song, Bo Zheng, et~al.
\newblock Longdocurl: a comprehensive multimodal long document benchmark integrating understanding, reasoning, and locating.
\newblock \emph{arXiv preprint arXiv:2412.18424}, 2024.

\bibitem[Ding et~al.(2022)Ding, Huang, Wang, Zhang, Chen, Ma, Chung, and Han]{ding2022v}
Yihao Ding, Zhe Huang, Runlin Wang, YanHang Zhang, Xianru Chen, Yuzhong Ma, Hyunsuk Chung, and Soyeon~Caren Han.
\newblock V-doc: Visual questions answers with documents.
\newblock In \emph{Proceedings of the IEEE/CVF conference on computer vision and pattern recognition}, pages 21492--21498, 2022.

\bibitem[Faysse et~al.(2024{\natexlab{a}})Faysse, Sibille, Wu, Omrani, Viaud, Hudelot, and Colombo]{faysse2024colpali}
Manuel Faysse, Hugues Sibille, Tony Wu, Bilel Omrani, Gautier Viaud, C{\'e}line Hudelot, and Pierre Colombo.
\newblock Colpali: Efficient document retrieval with vision language models.
\newblock In \emph{The Thirteenth International Conference on Learning Representations}, 2024{\natexlab{a}}.

\bibitem[Faysse et~al.(2024{\natexlab{b}})Faysse, Sibille, Wu, Omrani, Viaud, Hudelot, and Colombo]{faysse2024colpaliefficientdocumentretrieval}
Manuel Faysse, Hugues Sibille, Tony Wu, Bilel Omrani, Gautier Viaud, Céline Hudelot, and Pierre Colombo.
\newblock Colpali: Efficient document retrieval with vision language models, 2024{\natexlab{b}}.

\bibitem[Gao et~al.(2023)Gao, Xiong, Gao, Jia, Pan, Bi, Dai, Sun, Wang, and Wang]{gao2023retrieval}
Yunfan Gao, Yun Xiong, Xinyu Gao, Kangxiang Jia, Jinliu Pan, Yuxi Bi, Yi Dai, Jiawei Sun, Haofen Wang, and Haofen Wang.
\newblock Retrieval-augmented generation for large language models: A survey.
\newblock \emph{arXiv preprint arXiv:2312.10997}, 2, 2023.

\bibitem[Grattafiori et~al.(2024)Grattafiori, Dubey, Jauhri, Pandey, Kadian, Al-Dahle, Letman, Mathur, Schelten, Vaughan, et~al.]{grattafiori2024llama}
Aaron Grattafiori, Abhimanyu Dubey, Abhinav Jauhri, Abhinav Pandey, Abhishek Kadian, Ahmad Al-Dahle, Aiesha Letman, Akhil Mathur, Alan Schelten, Alex Vaughan, et~al.
\newblock The llama 3 herd of models.
\newblock \emph{arXiv preprint arXiv:2407.21783}, 2024.

\bibitem[Hu et~al.(2024)Hu, Xu, Ye, Yan, Zhang, Zhang, Li, Zhang, Jin, Huang, et~al.]{hu2024mplug}
Anwen Hu, Haiyang Xu, Jiabo Ye, Ming Yan, Liang Zhang, Bo Zhang, Chen Li, Ji Zhang, Qin Jin, Fei Huang, et~al.
\newblock mplug-docowl 1.5: Unified structure learning for ocr-free document understanding.
\newblock \emph{arXiv preprint arXiv:2403.12895}, 2024.

\bibitem[Hui et~al.(2024)Hui, Lu, and Zhang]{hui2024uda}
Yulong Hui, Yao Lu, and Huanchen Zhang.
\newblock Uda: A benchmark suite for retrieval augmented generation in real-world document analysis.
\newblock \emph{arXiv preprint arXiv:2406.15187}, 2024.

\bibitem[Kannan et~al.(2024)Kannan, Venkatesh, and Min]{kannan2024smart}
Shyam~Sundar Kannan, Vishnunandan~LN Venkatesh, and Byung-Cheol Min.
\newblock Smart-llm: Smart multi-agent robot task planning using large language models.
\newblock In \emph{2024 IEEE/RSJ International Conference on Intelligent Robots and Systems (IROS)}, pages 12140--12147. IEEE, 2024.

\bibitem[Khattab and Zaharia(2020)]{khattab2020colbert}
Omar Khattab and Matei Zaharia.
\newblock Colbert: Efficient and effective passage search via contextualized late interaction over bert.
\newblock In \emph{Proceedings of the 43rd International ACM SIGIR conference on research and development in Information Retrieval}, pages 39--48, 2020.

\bibitem[Kim et~al.(2024)Kim, Park, Jeong, Chan, Xu, McDuff, Lee, Ghassemi, Breazeal, Park, et~al.]{kim2024mdagents}
Yubin Kim, Chanwoo Park, Hyewon Jeong, Yik~Siu Chan, Xuhai Xu, Daniel McDuff, Hyeonhoon Lee, Marzyeh Ghassemi, Cynthia Breazeal, Hae Park, et~al.
\newblock Mdagents: An adaptive collaboration of llms for medical decision-making.
\newblock \emph{Advances in Neural Information Processing Systems}, 37:\penalty0 79410--79452, 2024.

\bibitem[Lewis et~al.(2020)Lewis, Perez, Piktus, Petroni, Karpukhin, Goyal, K{\"u}ttler, Lewis, Yih, Rockt{\"a}schel, et~al.]{lewis2020retrieval}
Patrick Lewis, Ethan Perez, Aleksandra Piktus, Fabio Petroni, Vladimir Karpukhin, Naman Goyal, Heinrich K{\"u}ttler, Mike Lewis, Wen-tau Yih, Tim Rockt{\"a}schel, et~al.
\newblock Retrieval-augmented generation for knowledge-intensive nlp tasks.
\newblock \emph{Advances in neural information processing systems}, 33:\penalty0 9459--9474, 2020.

\bibitem[Li et~al.(2024)Li, Zhang, Guo, Zhang, Li, Zhang, Zhang, Li, Liu, and Li]{li2024llava}
Bo Li, Yuanhan Zhang, Dong Guo, Renrui Zhang, Feng Li, Hao Zhang, Kaichen Zhang, Yanwei Li, Ziwei Liu, and Chunyuan Li.
\newblock Llava-onevision: Easy visual task transfer.
\newblock \emph{arXiv preprint arXiv:2408.03326}, 2024.

\bibitem[Li et~al.(2025)Li, Wang, Gu, Chang, and Peng]{li2025metal}
Bingxuan Li, Yiwei Wang, Jiuxiang Gu, Kai-Wei Chang, and Nanyun Peng.
\newblock Metal: A multi-agent framework for chart generation with test-time scaling.
\newblock \emph{arXiv preprint arXiv:2502.17651}, 2025.

\bibitem[Li et~al.(2023)Li, Hammoud, Itani, Khizbullin, and Ghanem]{li2023camel}
Guohao Li, Hasan Hammoud, Hani Itani, Dmitrii Khizbullin, and Bernard Ghanem.
\newblock Camel: Communicative agents for" mind" exploration of large language model society.
\newblock \emph{Advances in Neural Information Processing Systems}, 36:\penalty0 51991--52008, 2023.

\bibitem[Liu et~al.(2024{\natexlab{a}})Liu, Li, Li, and Lee]{liu2023improved}
Haotian Liu, Chunyuan Li, Yuheng Li, and Yong~Jae Lee.
\newblock Improved baselines with visual instruction tuning.
\newblock In \emph{Proceedings of the IEEE/CVF Conference on Computer Vision and Pattern Recognition}, pages 26296--26306, 2024{\natexlab{a}}.

\bibitem[Liu et~al.(2024{\natexlab{b}})Liu, Li, Wu, and Lee]{liu2023visual}
Haotian Liu, Chunyuan Li, Qingyang Wu, and Yong~Jae Lee.
\newblock Visual instruction tuning.
\newblock \emph{Advances in neural information processing systems}, 36, 2024{\natexlab{b}}.

\bibitem[Luo et~al.(2024)Luo, Shen, Zhu, Zheng, Yu, and Yao]{luo2024layoutllm}
Chuwei Luo, Yufan Shen, Zhaoqing Zhu, Qi Zheng, Zhi Yu, and Cong Yao.
\newblock Layoutllm: Layout instruction tuning with large language models for document understanding.
\newblock In \emph{Proceedings of the IEEE/CVF conference on computer vision and pattern recognition}, pages 15630--15640, 2024.

\bibitem[Ma et~al.(2024{\natexlab{a}})Ma, Zhuang, Koopman, Zuccon, Chen, and Lin]{ma2024visa}
Xueguang Ma, Shengyao Zhuang, Bevan Koopman, Guido Zuccon, Wenhu Chen, and Jimmy Lin.
\newblock Visa: Retrieval augmented generation with visual source attribution.
\newblock \emph{arXiv preprint arXiv:2412.14457}, 2024{\natexlab{a}}.

\bibitem[Ma et~al.(2024{\natexlab{b}})Ma, Zang, Chen, Chen, Jiao, Li, Lu, Liu, Ma, Dong, Zhang, Pan, Jiang, Wang, Cao, and Sun]{ma2024mmlongbenchdocbenchmarkinglongcontextdocument}
Yubo Ma, Yuhang Zang, Liangyu Chen, Meiqi Chen, Yizhu Jiao, Xinze Li, Xinyuan Lu, Ziyu Liu, Yan Ma, Xiaoyi Dong, Pan Zhang, Liangming Pan, Yu-Gang Jiang, Jiaqi Wang, Yixin Cao, and Aixin Sun.
\newblock Mmlongbench-doc: Benchmarking long-context document understanding with visualizations, 2024{\natexlab{b}}.

\bibitem[Marafioti et~al.(2025)Marafioti, Zohar, Farr\'{e}, Noyan, Bakouch, Cuenca, Zakka, Ben~Allal, Lozhkov, Tazi, Srivastav, Lochner, Larcher, Morlon, Tunstall, von Werra, and Wolf]{marafioti2025smolvlm}
Andr\'{e}s Marafioti, Orr Zohar, Miquel Farr\'{e}, Merve Noyan, Elie Bakouch, Pedro Cuenca, Cyril Zakka, Loubna Ben~Allal, Anton Lozhkov, Nouamane Tazi, Vaibhav Srivastav, Joshua Lochner, Hugo Larcher, Mathieu Morlon, Lewis Tunstall, Leandro von Werra, and Thomas Wolf.
\newblock Smolvlm: Redefining small and efficient multimodal models.
\newblock 2025.

\bibitem[Mishra et~al.(2019)Mishra, Shekhar, Singh, and Chakraborty]{mishra2019ocr}
Anand Mishra, Shashank Shekhar, Ajeet~Kumar Singh, and Anirban Chakraborty.
\newblock Ocr-vqa: Visual question answering by reading text in images.
\newblock In \emph{2019 international conference on document analysis and recognition (ICDAR)}, pages 947--952. IEEE, 2019.

\bibitem[Naveed et~al.(2023)Naveed, Khan, Qiu, Saqib, Anwar, Usman, Akhtar, Barnes, and Mian]{naveed2023comprehensive}
Humza Naveed, Asad~Ullah Khan, Shi Qiu, Muhammad Saqib, Saeed Anwar, Muhammad Usman, Naveed Akhtar, Nick Barnes, and Ajmal Mian.
\newblock A comprehensive overview of large language models.
\newblock \emph{arXiv preprint arXiv:2307.06435}, 2023.

\bibitem[OpenAI(2023)]{openai2023gpt4}
OpenAI.
\newblock Gpt-4 technical report, 2023.
\newblock \url{https://arxiv.org/abs/2303.08774}.

\bibitem[Park et~al.(2024)Park, Choi, Park, and Han]{park2024hierarchical}
Jaeyoo Park, Jin~Young Choi, Jeonghyung Park, and Bohyung Han.
\newblock Hierarchical visual feature aggregation for ocr-free document understanding.
\newblock \emph{Advances in Neural Information Processing Systems}, 37:\penalty0 105972--105996, 2024.

\bibitem[Santhanam et~al.(2021)Santhanam, Khattab, Saad-Falcon, Potts, and Zaharia]{santhanam2021colbertv2}
Keshav Santhanam, Omar Khattab, Jon Saad-Falcon, Christopher Potts, and Matei Zaharia.
\newblock Colbertv2: Effective and efficient retrieval via lightweight late interaction.
\newblock \emph{arXiv preprint arXiv:2112.01488}, 2021.

\bibitem[Su et~al.(2020)Su, Wang, Zeng, Tang, Chen, Qiu, and Wang]{su2020adapting}
Peng Su, Kun Wang, Xingyu Zeng, Shixiang Tang, Dapeng Chen, Di Qiu, and Xiaogang Wang.
\newblock Adapting object detectors with conditional domain normalization.
\newblock In \emph{Computer Vision--ECCV 2020: 16th European Conference, Glasgow, UK, August 23--28, 2020, Proceedings, Part XI 16}, pages 403--419. Springer, 2020.

\bibitem[Suri et~al.(2024)Suri, Mathur, Dernoncourt, Goswami, Rossi, and Manocha]{suri2024visdom}
Manan Suri, Puneet Mathur, Franck Dernoncourt, Kanika Goswami, Ryan~A Rossi, and Dinesh Manocha.
\newblock Visdom: Multi-document qa with visually rich elements using multimodal retrieval-augmented generation.
\newblock \emph{arXiv preprint arXiv:2412.10704}, 2024.

\bibitem[Tanaka et~al.(2023)Tanaka, Nishida, Nishida, Hasegawa, Saito, and Saito]{tanaka2023slidevqa}
Ryota Tanaka, Kyosuke Nishida, Kosuke Nishida, Taku Hasegawa, Itsumi Saito, and Kuniko Saito.
\newblock Slidevqa: A dataset for document visual question answering on multiple images.
\newblock In \emph{Proceedings of the AAAI Conference on Artificial Intelligence}, pages 13636--13645, 2023.

\bibitem[Tito et~al.(2023)Tito, Karatzas, and Valveny]{tito2023hierarchical}
Rub{\`e}n Tito, Dimosthenis Karatzas, and Ernest Valveny.
\newblock Hierarchical multimodal transformers for multipage docvqa.
\newblock \emph{Pattern Recognition}, 144:\penalty0 109834, 2023.

\bibitem[Tong et~al.(2025)Tong, Wang, Chen, Ji, Qiu, Han, Geng, Xue, Zhou, Xia, et~al.]{tong2025mj}
Haibo Tong, Zhaoyang Wang, Zhaorun Chen, Haonian Ji, Shi Qiu, Siwei Han, Kexin Geng, Zhongkai Xue, Yiyang Zhou, Peng Xia, et~al.
\newblock Mj-video: Fine-grained benchmarking and rewarding video preferences in video generation.
\newblock \emph{arXiv preprint arXiv:2502.01719}, 2025.

\bibitem[Wang et~al.(2024)Wang, Bai, Tan, Wang, Fan, Bai, Chen, Liu, Wang, Ge, Fan, Dang, Du, Ren, Men, Liu, Zhou, Zhou, and Lin]{Qwen2-VL}
Peng Wang, Shuai Bai, Sinan Tan, Shijie Wang, Zhihao Fan, Jinze Bai, Keqin Chen, Xuejing Liu, Jialin Wang, Wenbin Ge, Yang Fan, Kai Dang, Mengfei Du, Xuancheng Ren, Rui Men, Dayiheng Liu, Chang Zhou, Jingren Zhou, and Junyang Lin.
\newblock Qwen2-vl: Enhancing vision-language model's perception of the world at any resolution.
\newblock \emph{arXiv preprint arXiv:2409.12191}, 2024.

\bibitem[Wu et~al.(2023)Wu, Bansal, Zhang, Wu, Li, Zhu, Jiang, Zhang, Zhang, Liu, et~al.]{wu2023autogen}
Qingyun Wu, Gagan Bansal, Jieyu Zhang, Yiran Wu, Beibin Li, Erkang Zhu, Li Jiang, Xiaoyun Zhang, Shaokun Zhang, Jiale Liu, et~al.
\newblock Autogen: Enabling next-gen llm applications via multi-agent conversation.
\newblock \emph{arXiv preprint arXiv:2308.08155}, 2023.

\bibitem[Xia et~al.(2024{\natexlab{a}})Xia, Chen, Tian, Gong, Hou, Xu, Wu, Fan, Zhou, Zhu, et~al.]{xia2024cares}
Peng Xia, Ze Chen, Juanxi Tian, Yangrui Gong, Ruibo Hou, Yue Xu, Zhenbang Wu, Zhiyuan Fan, Yiyang Zhou, Kangyu Zhu, et~al.
\newblock Cares: A comprehensive benchmark of trustworthiness in medical vision language models.
\newblock \emph{Advances in Neural Information Processing Systems}, 37:\penalty0 140334--140365, 2024{\natexlab{a}}.

\bibitem[Xia et~al.(2024{\natexlab{b}})Xia, Han, Qiu, Zhou, Wang, Zheng, Chen, Cui, Ding, Li, et~al.]{xia2024mmie}
Peng Xia, Siwei Han, Shi Qiu, Yiyang Zhou, Zhaoyang Wang, Wenhao Zheng, Zhaorun Chen, Chenhang Cui, Mingyu Ding, Linjie Li, et~al.
\newblock Mmie: Massive multimodal interleaved comprehension benchmark for large vision-language models.
\newblock \emph{arXiv preprint arXiv:2410.10139}, 2024{\natexlab{b}}.

\bibitem[Xia et~al.(2024{\natexlab{c}})Xia, Zhu, Li, Wang, Shi, Wang, Zhang, Zou, and Yao]{xia2024mmed}
Peng Xia, Kangyu Zhu, Haoran Li, Tianze Wang, Weijia Shi, Sheng Wang, Linjun Zhang, James Zou, and Huaxiu Yao.
\newblock Mmed-rag: Versatile multimodal rag system for medical vision language models.
\newblock \emph{arXiv preprint arXiv:2410.13085}, 2024{\natexlab{c}}.

\bibitem[Xia et~al.(2024{\natexlab{d}})Xia, Zhu, Li, Zhu, Li, Li, Zhang, and Yao]{xia2024rule}
Peng Xia, Kangyu Zhu, Haoran Li, Hongtu Zhu, Yun Li, Gang Li, Linjun Zhang, and Huaxiu Yao.
\newblock Rule: Reliable multimodal rag for factuality in medical vision language models.
\newblock In \emph{Proceedings of the 2024 Conference on Empirical Methods in Natural Language Processing}, pages 1081--1093, 2024{\natexlab{d}}.

\bibitem[Xing et~al.(2025)Xing, Wang, Li, Bai, Wang, Qian, Yao, and Tu]{xing2025re}
Shuo Xing, Yuping Wang, Peiran Li, Ruizheng Bai, Yueqi Wang, Chengxuan Qian, Huaxiu Yao, and Zhengzhong Tu.
\newblock Re-align: Aligning vision language models via retrieval-augmented direct preference optimization.
\newblock \emph{arXiv preprint arXiv:2502.13146}, 2025.

\bibitem[Zhang et~al.(2024{\natexlab{a}})Zhang, Zhang, Wang, Ouyang, Wen, Li, Chow, He, and Zhang]{zhang2024ocr}
Junyuan Zhang, Qintong Zhang, Bin Wang, Linke Ouyang, Zichen Wen, Ying Li, Ka-Ho Chow, Conghui He, and Wentao Zhang.
\newblock Ocr hinders rag: Evaluating the cascading impact of ocr on retrieval-augmented generation.
\newblock \emph{arXiv preprint arXiv:2412.02592}, 2024{\natexlab{a}}.

\bibitem[Zhang et~al.(2024{\natexlab{b}})Zhang, Huang, Liu, Tang, Lu, Chen, Bai, Chu, Yu, and Ouyang]{zhang2024motiongpt}
Yaqi Zhang, Di Huang, Bin Liu, Shixiang Tang, Yan Lu, Lu Chen, Lei Bai, Qi Chu, Nenghai Yu, and Wanli Ouyang.
\newblock Motiongpt: Finetuned llms are general-purpose motion generators.
\newblock In \emph{Proceedings of the AAAI Conference on Artificial Intelligence}, pages 7368--7376, 2024{\natexlab{b}}.

\bibitem[Zhou et~al.(2023)Zhou, Cui, Yoon, Zhang, Deng, Finn, Bansal, and Yao]{zhou2023analyzing}
Yiyang Zhou, Chenhang Cui, Jaehong Yoon, Linjun Zhang, Zhun Deng, Chelsea Finn, Mohit Bansal, and Huaxiu Yao.
\newblock Analyzing and mitigating object hallucination in large vision-language models.
\newblock \emph{arXiv preprint arXiv:2310.00754}, 2023.

\bibitem[Zhou et~al.(2024{\natexlab{a}})Zhou, Cui, Rafailov, Finn, and Yao]{zhou2024aligning}
Yiyang Zhou, Chenhang Cui, Rafael Rafailov, Chelsea Finn, and Huaxiu Yao.
\newblock Aligning modalities in vision large language models via preference fine-tuning.
\newblock \emph{arXiv preprint arXiv:2402.11411}, 2024{\natexlab{a}}.

\bibitem[Zhou et~al.(2024{\natexlab{b}})Zhou, Fan, Cheng, Yang, Chen, Cui, Wang, Li, Zhang, and Yao]{zhou2024calibrated}
Yiyang Zhou, Zhiyuan Fan, Dongjie Cheng, Sihan Yang, Zhaorun Chen, Chenhang Cui, Xiyao Wang, Yun Li, Linjun Zhang, and Huaxiu Yao.
\newblock Calibrated self-rewarding vision language models.
\newblock \emph{arXiv preprint arXiv:2405.14622}, 2024{\natexlab{b}}.

\bibitem[Zhu et~al.(2023)Zhu, Chen, Shen, Li, and Elhoseiny]{zhu2023minigpt}
Deyao Zhu, Jun Chen, Xiaoqian Shen, Xiang Li, and Mohamed Elhoseiny.
\newblock Minigpt-4: Enhancing vision-language understanding with advanced large language models.
\newblock \emph{arXiv preprint arXiv:2304.10592}, 2023.

\bibitem[Zhu et~al.(2024)Zhu, Xia, Li, Zhu, Wang, and Yao]{zhu2024mmedpo}
Kangyu Zhu, Peng Xia, Yun Li, Hongtu Zhu, Sheng Wang, and Huaxiu Yao.
\newblock Mmedpo: Aligning medical vision-language models with clinical-aware multimodal preference optimization.
\newblock \emph{arXiv preprint arXiv:2412.06141}, 2024.

\end{thebibliography}
}

\ifarxiv \clearpage \appendix \section{Experimental Setup}
\label{sec:appendix_section}

\subsection{Baseline Models}
\begin{itemize}
    \item \textbf{Qwen2-VL-7B-Instruct}~\cite{Qwen2-VL}: A large vision-language model developed by Alibaba, designed to handle multiple images as input. 
    \item \textbf{Qwen2.5-VL-7B-Instruct}~\cite{Qwen2.5-VL}: An enhanced version of Qwen2-VL-7B-Instruct, offering improved performance in processing multiple images. 
    \item \textbf{llava-v1.6-mistral-7b}~\cite{liu2023improved}: Also called LLaVA-NeXT, a vision-language model improved upon LLaVa-1.5, capable of interpreting and generating content from multiple images. 
    \item \textbf{Phi-3.5-vision-instruct}~\cite{abdin2024phi}: A model developed by Microsoft that integrates vision and language understanding, designed to process and generate responses based on multiple images. 
    \item \textbf{llava-one-vision-7B}~\cite{li2024llava}: A model trained on LLaVA-OneVision, based on Qwen2-7B language model with a context window of 32K tokens.
    \item \textbf{SmolVLM-Instruct}~\cite{marafioti2025smolvlm}: A compact vision-language model developed by HuggingFace, optimized for handling image inputs efficiently.
    \item \textbf{ColBERTv2+Llama-3.1-8B-Instruct}~\cite{santhanam2021colbertv2,grattafiori2024llama}: A text-based RAG pipeline that utilizes ColBERTv2~\cite{santhanam2021colbertv2} for retrieving text segments and Llama-3.1-8B-Instruct as the LLM to generate responses.
    \item \textbf{M3DocRAG}~\cite{cho2024m3docrag}: An image-based RAG pipeline that employs ColPali~\cite{faysse2024colpali} for retrieving image segments and Qwen2-VL-7B-Instruct~\cite{Qwen2-VL} as the LVLM for answer generation.
\end{itemize}

\begin{table*}[htbp]
\centering
\renewcommand{\arraystretch}{1.2}
\setlength{\tabcolsep}{5pt}
\begin{tabular}{lcccccc}
\toprule
\textbf{Method} & \textbf{Layout} & \textbf{Text} & \textbf{Figure} & \textbf{Table} & \textbf{Others} & \textbf{Avg} \\
\midrule
\multicolumn{7}{c}{\textit{LVLMs}} \\
\midrule
Qwen2-VL-7B-Instruct & 0.264 & 0.386 & 0.308 & 0.207 & 0.500 & 0.296 \\
Qwen2.5-VL-7B-Instruct & 0.357 & 0.479 & 0.442 & 0.299 & 0.375 & 0.389 \\
llava-v1.6-mistral-7b & 0.067 & 0.165 & 0.088 & 0.051 & 0.250 & 0.099 \\
llava-one-vision-7B & 0.098 & 0.200 & 0.144 & 0.057 & 0.125 & 0.126 \\
Phi-3.5-vision-instruct & 0.245 & 0.375 & 0.291 & 0.187 & 0.375 & 0.280 \\
SmolVLM-Instruct & 0.128 & 0.224 & 0.164 & 0.100 & 0.250 & 0.163 \\
\midrule
\multicolumn{7}{c}{\textit{RAG methods (top 1)}} \\
\midrule
ColBERTv2+Llama-3.1-8B & 0.257 & 0.529 & 0.471 & 0.428 & \textbf{0.775} & 0.429 \\
M3DocRAG (ColPali+Qwen2-VL-7B) & 0.340 & 0.605 & \textbf{0.546} & 0.520 & 0.625 & 0.506 \\
\textbf{\ours\ (Ours)} & \textbf{0.341} & \textbf{0.612} & 0.540 & \textbf{0.527} & 0.750 & \textbf{0.517} \\
\midrule
\multicolumn{7}{c}{\textit{RAG methods (top 4)}} \\
\midrule
ColBERTv2+Llama-3.1-8B & 0.349 & 0.599 & 0.491 & 0.485 & \textbf{0.875} & 0.491 \\
M3DocRAG (ColPali+Qwen2-VL-7B) & 0.426 & 0.660 & 0.595 & 0.542 & 0.625 & 0.554 \\
\textbf{\ours\ (Ours)} & \textbf{0.438} & \textbf{0.675} & \textbf{0.592} & \textbf{0.581} & \textbf{0.875} & \textbf{0.578} \\
\bottomrule
\end{tabular}
\caption{Performance comparison across different evidence source on LongDocURL.}
\label{tab:ldu-results}
\end{table*}

\subsection{Evaluation Benchmarks}
\label{sec:appendix_datainfo}
\begin{itemize}
    \item \textbf{MMLongBench}~\cite{ma2024mmlongbenchdocbenchmarkinglongcontextdocument}: Evaluates models' ability to understand long documents with rich layouts and multi-modal components, comprising 1091 questions and 135 documents averaging 47.5 pages each. 
    \item \textbf{LongDocURL}~\cite{deng2024longdocurl}: Provides a comprehensive multi-modal long document benchmark integrating understanding, reasoning, and locating tasks, covering over 33,000 pages of documents and 2,325 question-answer pairs. 
    \item \textbf{PaperTab}~\cite{hui2024uda}: Focuses on evaluating models' ability to comprehend and extract information from tables within NLP research papers, covering 393 questions among 307 documents.
    \item \textbf{PaperText}~\cite{hui2024uda}: Assesses models' proficiency in understanding the textual content of NLP research papers, covering 2804 questions among 1087 documents.
    \item \textbf{FetaTab}~\cite{hui2024uda}: a question-answering dataset for tables from Wikipedia pages, challengeing models to generate free-form text answers, comprising 1023 questions and 878 documents.
\end{itemize}

\subsection{Hyperparameter Settings}

\begin{itemize}
    \item \textbf{Temperature}: All models use their default temperature setting.
    \item \textbf{Max New Tokens}: 256.
    \item \textbf{Max Tokens per Image (Qwen2-VL-7B-Instruct)}:  
    \begin{itemize}
        \item \textbf{Top-1 retrieval}: 16,384 (by default).
        \item \textbf{Top-4 retrieval}: 2,048.
    \end{itemize}
    \item \textbf{Image Resolution}: 144 (for all benchmarks).
\end{itemize}

\subsection{Prompt Settings}
\label{sec:appendix_prompt_section}

\begin{tcolorbox}[title=General Agent]
You are an advanced agent capable of analyzing both text and images. Your task is to use both the textual and visual information provided to answer the user’s question accurately.

\textbf{Extract Text from Both Sources}: If the image contains text, extract it and consider both the text in the image and the provided textual content.  

\textbf{Analyze Visual and Textual Information}: Combine details from both the image (e.g., objects, scenes, or patterns) and the text to build a comprehensive understanding of the content.

\textbf{Provide a Combined Answer}: Use the relevant details from both the image and the text to provide a clear, accurate, and context-aware response to the user's question.

\textbf{When responding:}
\begin{itemize}
    \item If both the image and text contain similar or overlapping information, cross-check and use both to ensure consistency.
    \item If the image contains information not present in the text, include it in your response if it is relevant to the question.
    \item If the text and image offer conflicting details, explain the discrepancies and clarify the most reliable source.
\end{itemize}
\end{tcolorbox}

\begin{tcolorbox}[title=Critical Agent]
Provide a Python dictionary of critical information based on all given information—one for text and one for image.

Respond exclusively in a valid dictionary format without any additional text. The format should be:

\{"text": "critical information for text", "image": "critical information for image"\}
\end{tcolorbox}

\begin{tcolorbox}[title=Text Agent]
You are a text analysis agent. Your job is to extract key information from the text and use it to answer the user’s question accurately.

\textbf{Your tasks:}
\begin{itemize}
    \item Extract key details. Focus on the most important facts, data, or ideas related to the question.
    \item Understand the context and pay attention to the meaning and details.
    \item Use the extracted information to give a concise and relevant response to the user's question. Provide a clear answer.
\end{itemize}
\end{tcolorbox}

\begin{tcolorbox}[title=Image Agent]
You are an advanced image processing agent specialized in analyzing and extracting information from images. The images may include document screenshots, illustrations, or photographs.

\textbf{Your tasks:}
\begin{itemize}
    \item Extract textual information from images using Optical Character Recognition (OCR).
    \item Analyze visual content to identify relevant details (e.g., objects, patterns, scenes).
    \item Combine textual and visual information to provide an accurate and context-aware answer to the user's question.
\end{itemize}
\end{tcolorbox}

\begin{tcolorbox}[title=Summarizing Agent]
You are tasked with summarizing and evaluating the collective responses provided by multiple agents. You have access to the following information:
\begin{itemize}
    \item \textbf{Answers}: The individual answers from all agents.
\end{itemize}

\textbf{Your tasks:}
\begin{itemize}
    \item \textbf{Analyze}: Evaluate the quality, consistency, and relevance of each answer. Identify commonalities, discrepancies, or gaps in reasoning.
    \item \textbf{Synthesize}: Summarize the most accurate and reliable information based on the evidence provided by the agents and their discussions.
    \item \textbf{Conclude}: Provide a final, well-reasoned answer to the question or task. Your conclusion should reflect the consensus (if one exists) or the most credible and well-supported answer.
\end{itemize}

Return the final answer in the following dictionary format:

\{"Answer": Your final answer here\}
\end{tcolorbox}

\begin{tcolorbox}[title=Evaluation]
\textbf{Question}: \{question\}  

\textbf{Predicted Answer}: \{answer\}  

\textbf{Ground Truth Answer}: \{gt\}  

Please evaluate whether the predicted answer is correct.  

\begin{itemize}
    \item If the answer is correct, return 1.
    \item If the answer is incorrect, return 0.
\end{itemize}

Return only a string formatted as a valid JSON dictionary that can be parsed using \texttt{json.loads}, for example: \{"correctness": 1\}
\end{tcolorbox}

\subsection{Evaluation Metrics}
The metric of all benchmarks is the average binary correctness evaluated by GPT-4o. The evaluation prompt is given in Section \ref{sec:appendix_prompt_section}. We use a python script to extract the result provided by GPT-4o.

\section{Additional Results}

\subsection{Fine-grained Performance of LongDocURL}
\label{sec:appendix_ldu}
We present the fine-grained performance of LongDocURL, as illustrated in Table~\ref{tab:ldu-results}. Similar to MMLongBench, \ours\ outperforms all LVLM baselines. When using the top 1 retrieval approach, though M3DocRAG performs slightly better on Figure and ColBERTv2+Llama3.1-8B performs slightly better on the type Others, \ours\ show strong performance in Layout, Text, Table and get the highest average accuracy. With the top 4 retrieval strategy, \ours\ improves its performance and reach the highest score in the all categories.

\subsection{Experiments on different model backbones in \ours}
\label{sec:appendix_more_lvlms}

\begin{table*}[ht]
    \centering
    \resizebox{0.85\textwidth}{!}{
    \setlength{\tabcolsep}{8pt}
    \begin{tabular}{l|cccccc}
        \toprule
         & \textbf{MMLongBench} & \textbf{LongDocUrl} & \textbf{PaperTab} & \textbf{PaperText} & \textbf{FetaTab} & \textbf{Avg} \\
        \midrule
        \multicolumn{7}{c}{\textit{With top 1 retrieval}} \\
        \midrule
        \textbf{+Qwen2-VL-7B-Instruct} & 0.299 & 0.517 & 0.219 & 0.399 & 0.600 & 0.407 \\
        \textbf{+Qwen2.5-VL-7B-Instruct} & 0.351 & 0.519 & 0.211 & 0.382 & 0.589 & 0.410 \\
        \textbf{+GPT-4o~\cite{openai2023gpt4}} & \textbf{0.420} & \textbf{0.595} & \textbf{0.293} & \textbf{0.474} & \textbf{0.716} & \textbf{0.500} \\
        \midrule
        \multicolumn{7}{c}{\textit{With top 4 retrieval}} \\
        \midrule
        \textbf{+Qwen2-VL-7B-Instruct} & 0.315 & \textbf{0.578} & \textbf{0.278} & \textbf{0.487} & \textbf{0.675} & 0.467 \\
        \textbf{+Qwen2.5-VL-7B-Instruct} & \textbf{0.389} & 0.566 & 0.277 & 0.454 & 0.671 & \textbf{0.471} \\
        \bottomrule
    \end{tabular}
    }
    \caption{Performance comparison of using different backbone LVLMs in \ours.}
    \label{tab:vlm_model_comparison}
\end{table*}
Table \ref{tab:vlm_model_comparison} presents an ablation study evaluating the impact of different LVLMs on the performance of our framework. Three LVLMs: Qwen2-VL-7B-Instruct, Qwen2.5-VL-7B-Instruct, and GPT-4o were integrated as the backbone model for all agents except the text agent.

Qwen2.5-VL-7B-Instruct performs worse than Qwen2-VL-7B-Instruct on PaperTab, PaperText, and FetaTab, with both top-1 and top-4 retrieval. However, Qwen2.5-VL shows an extremely marked improvement over Qwen2-VL on MMLongBench, resulting higher average scores. MMLongBench's greater reliance on image-based questions might explain Qwen2.5-VL's superior performance on this benchmark, possibly indicating that Qwen2.5-VL is better at handling visual question-answering tasks, but worse at handling textual tasks.

Importantly, GPT-4o significantly outperforms both Qwen2-VL and Qwen2.5-VL across all benchmarks. Remarkably, GPT-4o's top-1 performance surpasses even the top-4 results of both Qwen models in almost all cases. This substantial performance increase strongly suggests that our framework effectively leverages more powerful backbone models, showcasing its adaptability and capacity to benefit from improvements in the underlying LVLMs.

\subsection{Additional case studies}
\label{sec:appendix_cases}

In Figure \ref{fig:case_study2}, the question requires identifying a reason from a list that lacks explicit numbering and is accompanied by images. ColBERT fails to retrieve the correct evidence page, resulting ColBERT + Llama's inability to answer the question. Although ColPali correctly locates the evidence page, M3DocRAG fails to get the correct answer. However, our framework successfully identifies the correct answer ("Most Beautiful Campus") through the concerted efforts of all agents. The general agent arrives at a preliminary answer and the critical agent identifies critical textual clues ("Most Beautiful Campus") and corresponding visual elements (images of the NTU campus). Image agent then refines the answer, leveraging the critical information to correctly pinpoint the description lacking people. Though text agent can't find the related information from the given context, information provided by the critical agent helps it to guess that the answer is "Most Beautiful Campus". The summarizing agent combines these insights to arrive at the correct final answer.

In Figure \ref{fig:case_study3}, the question asks for Professor Lebour's degree. ColPali fails to retrieve the relevant page, rendering M3DocRAG ineffective. While ColBERT correctly retrieves the page, ColBERT + Llama still produces an incorrect answer because it incorrectly adds "F.G.S." to the answer, which is not a degree. \ours, on the other hand, correctly identifies the "M.A. degree". The general agent provides an initial answer, and the critical agent identifies the "M.A." designation in both text and image. Based on the clue, the text agent adds a more detailed explanation, and the image agent directly uses the clue as its answer. Finally, the summarizing agent synthesizes the results to provide the verified answer.

These two cases highlight \ours's resilience to imperfect retrieval, demonstrating the effectiveness of collaborative multi-modal information processing and the importance of the general-critical agent's guidance in achieving high accuracy even with potentially insufficient or ambiguous information.

\begin{figure*}[tp]
    \centering
    \includegraphics[width=0.98\textwidth, height=0.4\textheight, keepaspectratio]{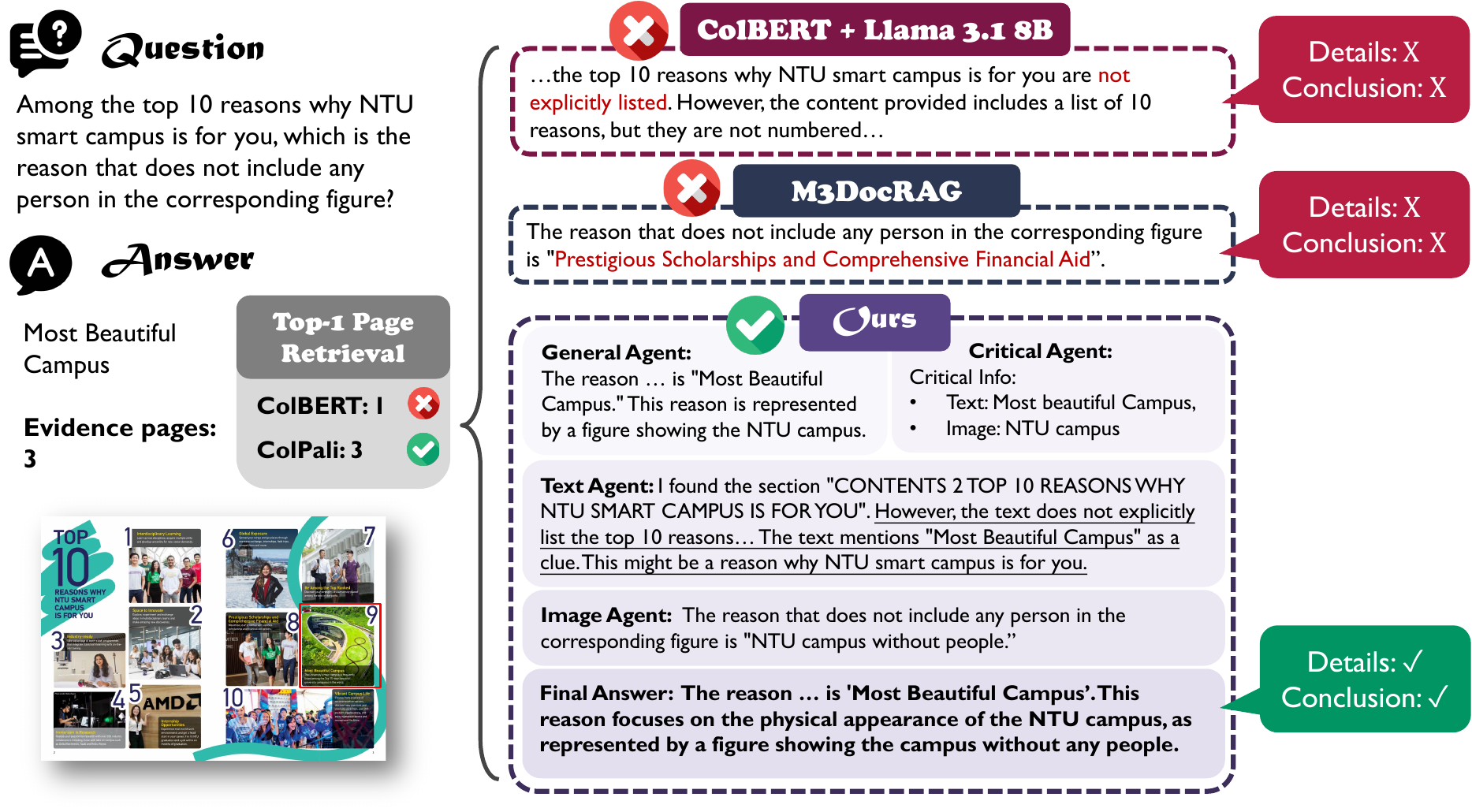}
    \caption{A Case study of \textbf{\ours} compared with other two baselines. While only ColPali correctly retrieves the evidence page, neither baseline method identifies the correct answer. Our method, through critical information sharing and specialized agent collaboration, correctly pinpoints the "Most Beautiful Campus" as the only reason without a corresponding image containing people.}
    \label{fig:case_study2}
    \vspace{-1.5em}
\end{figure*}
\begin{figure*}[tp]
    \centering
    \includegraphics[width=0.98\textwidth, height=0.4\textheight, keepaspectratio]{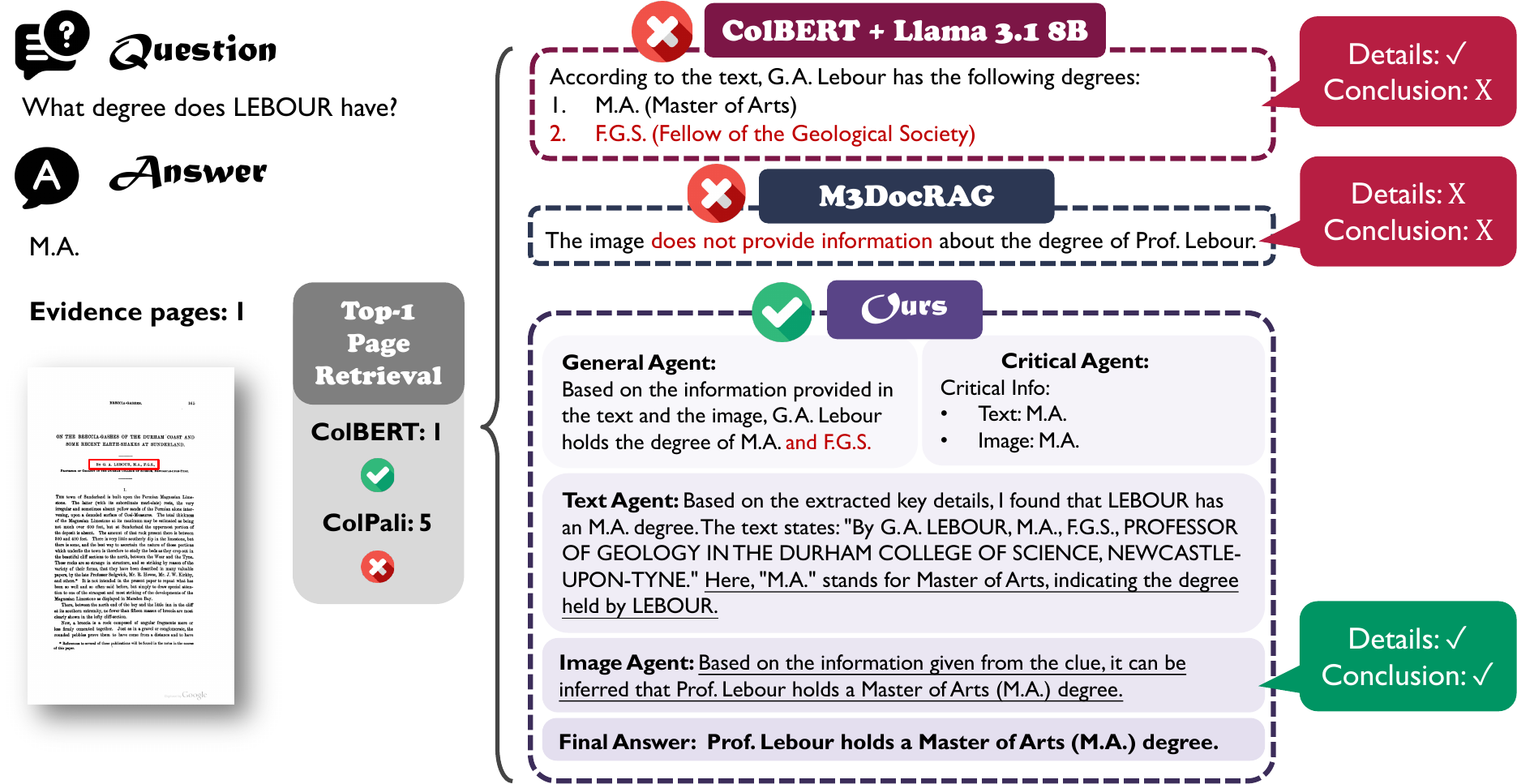}
    \caption{A Case study of \textbf{\ours} compared with other two RAG-method baselines. In this case, ColPali fails to retrieve the correct evidence page, hindering M3DocRAG. While ColBERT succeeds in retrieval, the ColBERT + Llama baseline still provides an incorrect answer. Only our multi-agent framework, through precise critical information extraction and agent collaboration, correctly identifies the M.A. degree.}
    \label{fig:case_study3}
    \vspace{-1.5em}
\end{figure*} \fi

\end{document}